# Beyond the Known: Adversarial Autoencoders in Novelty Detection

Muhammad Asad[1,2], Ihsan Ullah[1,2], Ganesh Sistu[3], Michael G. Madden[1,2]
[1]*Machine Learning Research Group, School of Computer Science, University of Galway, Ireland*
[2]*Insight SFI Research Centre for Data Analytics, University of Galway, Ireland*
[3] *Valeo Vision Systems, Tuam, Ireland*
{*m.asad2, ihsan.ullah, michael.madden*}*@universityofgalway.ie, ganesh.sistu@valeo.com*

Keywords: Anomaly Detection, Adversarial Auto-encoders, Reconstruction, Discriminator, Distribution

Abstract: In novelty detection, the goal is to decide if a new data point should be categorized as an inlier or an outlier, given a training dataset that primarily captures the inlier distribution. Recent approaches typically use deep encoder and decoder network frameworks to derive a reconstruction error, and employ this error either to determine a novelty score, or as the basis for a one-class classifier. In this research, we use a similar framework but with a lightweight deep network, and we adopt a probabilistic score with reconstruction error. Our methodology calculates the probability of whether the sample comes from the inlier distribution or not. This work makes two key contributions. The first is that we compute the novelty probability by linearizing the manifold that holds the structure of the inlier distribution. This allows us to interpret how the probability is distributed and can be determined in relation to the local coordinates of the manifold tangent space. The second contribution is that we improve the training protocol for the network. Our results indicate that our approach is effective at learning the target class, and it outperforms recent state-of-the-art methods on several benchmark datasets.

## 1 INTRODUCTION

Novelty detection is one of the main challenges in computer vision data analysis. Its aim is to distinguish whether a new data point aligns with typical patterns (is an inlier) or deviates from them (is an outlier) (Almohsen et al., 2022). This task is challenging because, while the inlier class is usually well characterized, there is generally insufficient information about the distribution of outliers. A related challenge is that outliers are typically infrequent and in some cases almost never observed, as seen in cases such as industrial fault detection (Liu et al., 2018). The significance of this methodology is not just confined to one domain; it arises in many applications, from medical diagnostics and drug discovery to computer vision tasks (Pidhorskyi et al., 2018) such as anomaly detection in images and videos. In these, it is important to detect new or unexplained data points and to respond to outliers. Detecting these outliers ensures that models are efficient and can learn the variations in the data.

In the area of computer vision, for instance, novelty detection is instrumental for detecting outliers (You et al., 2017), denoising images, and finding anomalies in visual media. Authors see novelty detection through the approach of one-class classification (Sabokrou et al., 2018), an approach that works well when negative classes are absent, less in number, or uncertain. In this context, the negative class is a novelty, outlier, or anomaly, while the positive class is well characterized in the training data instances (Bergadano, 2019; Zhang et al., 2016).

Modern novelty detection approaches are leveraging the capabilities of deep learning. These advanced techniques often take one of two paths: they either develop a one-class classifier (Almohsen et al., 2022) or utilize the reconstruction in encoder-decoder models to deal with novelty (Ravanbakhsh et al., 2017).

In this research, we present an autoencoder architecture inspired by adversarial autoencoders. Unlike existing methods that train a one-class classifier (Almohsen et al., 2022), we focus on understanding the probability distribution of inlier data points. This process simplifies the task of novelty detection by checking if those samples are less common or potential outliers. These samples are then confirmed as outliers if they fall below a certain threshold, as discussed in Pidhorskyi et al. (2018). In recent years, the work by Pidhorskyi et al. (2018) has made a significant contribution to the field of novelty detection. This methodology has been adopted widely (Zhou, 2022) because of the manifold learning that captures the distribution structure of inliers and by determining if a specific sample is an anomaly by looking at its likelihood distribution.

Building upon the work in Pidhorskyi et al. (2018), we have developed a method in which we determine the probability distribution of the entire model, which can cover both the signal and the noise. Our main goal is novelty detection in images and managing the latent space distribution by ensuring that it can accurately represent the inlier distribution. This is not just about generating good images; it is also about getting an accurate novelty score. Many studies in deep learning have focused on the reconstruction error (Ionescu et al., 2019; Xia et al., 2015; Pidhorskyi et al., 2018). We use that too, but in our system, the reconstruction error is mainly related to the noise from the reconstruction of the outliers. After getting the latent distribution and improving the image generation, we use an adversarial autoencoder network with two discriminators. These discriminators help us tackle both challenges.

We named our approach *Beyond the Known: Adversarial Autoencoders in Novelty Detection* as BK-AAND. Our method has a unique advantage it allows the decoder network to effectively learn and map out the shape of the inlier distribution. This is done by understanding the probability distribution of the latent space. What makes our approach efficient is how we handle the manifold for a given test sample. We make it linear and show that, based on local manifold coordinates, the data distribution splits into two parts. One part is influenced by the manifold itself, which includes the decoder network and latent distribution. The other part is influenced by the reconstruction error. While we do take advantage of that, our framework shows that the reconstruction error only influences the noise part of the model.

Our main contributions include:

- Introducing a lightweight approach based on computing the loss and probability distribution for both inliers and the entire model (Section 3).

- Evaluating our model on a variety of datasets, each with different outlier percentages, ensuring its performance across different scenarios.

- Performing a comparative analysis with state-of-the-art techniques in the field (Section 5), using AUC and $F_1$ score.

In Section 2, related studies and literature are discussed. Section 3 outlines our methodology and the framework. Section 4 details the experiments we conducted. The results we obtained are discussed in Section 5. Our final thoughts and conclusions based on the findings are presented in Section 6.

## 2 RELATED WORK

Anomaly detection can be addressed as a novelty detection problem, as done in many research studies (Liu et al., 2018; Zhang et al., 2016; Luo et al., 2017; Hinami et al., 2017; Xia et al., 2015; Sultani et al., 2018; Sabokrou et al., 2018; Bergadano, 2019; Hasan et al., 2016; Smeureanu et al., 2017; Ravanbakhsh et al., 2018, 2017). The goal is to train a model on recognized normal data and then identify unknown data/outliers as anomalies. With the rise of deep learning approaches, there have been suggestions (Smeureanu et al., 2017; Ravanbakhsh et al., 2017) to use the pre-trained convolutional network features for the training of one-class classifiers. The effectiveness of these approaches mostly relies on the foundational model, which is frequently trained on unrelated datasets (Sabokrou et al., 2018).

More recent developments in this area use generative networks to learn features, as discussed in various studies (Liu et al., 2023; Zhou and Xing, 2023; Gong et al., 2019; Ren et al., 2015; Xu et al., 2015; Ionescu et al., 2019; Xu et al., 2017; Sabokrou et al., 2018). Specifically, Ionescu et al. (2019) suggested using convolutional auto-encoders that are combined with object detection to learn both motion and appearance representations. Several studies (Yamanishi et al., 2000; Humbert et al., 2022; Jiang et al., 2022) utilize statistical techniques to better understand and capture the common patterns within data. They create models based on these patterns and anything that deviates significantly from these models is considered unusual or as an outlier. Essentially, these methods learn from the usual trends and patterns in the data and then point out the rare occurrences that do not follow the trend.

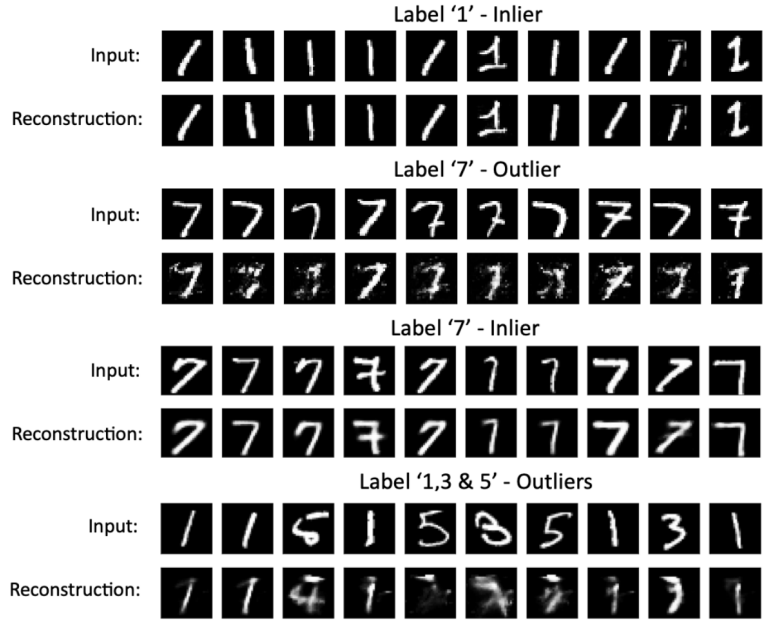

Figure 1: Illustration of reconstructions of inliers and outliers on MNIST images. The figure displays the reconstructions generated by the autoencoder network, which was initially trained on data labeled "1" in MNIST (LeCun, 1998). The first row represents the input images of the inliers with the label '1', while the second row shows their respective reconstructions. Meanwhile, the third row displays input images of outliers with the label '7', and the fourth row presents the corresponding reconstructions and so on. For a fair comparison, similar digits are considered.

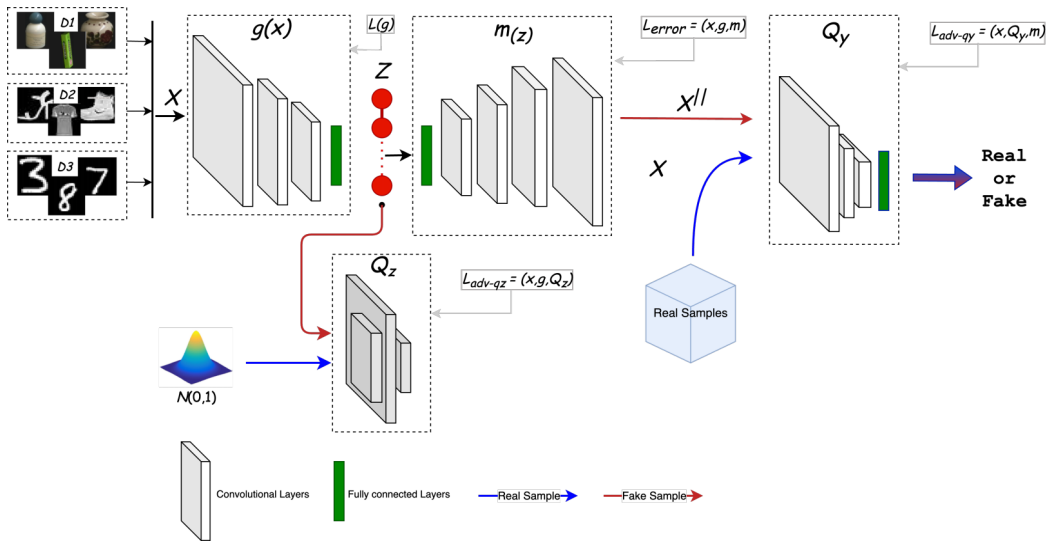

Figure 2: Framework of the BK-AAND network. The network is designed to train an Adversarial Autoencoder (AAE) (Sabokrou et al., 2018, 2017). In line with previous works (Almohsen et al., 2022; Pidhorskyi et al., 2018), it adds an additional adversarial component that enhances the generative capabilities of images that are decoded and then improves the manifold learning.

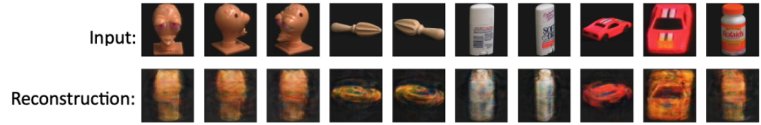

Figure 3: Illustration of reconstruction of outliers on Coil-100 images. The figure displays the reconstructions generated by the autoencoder network, which was initially trained on one image. The first row represents the input images of the outliers, while the second row shows their respective reconstructions.

Some authors (Xu et al., 2022; Angiulli et al., 2020; Wang et al., 2022) use a different approach,

instead of modeling the usual patterns, they look at the distances between data points. They believe that normal data points tend to cluster together, while outliers are more isolated, maintaining greater distances from these clusters. The LOF method (Breunig et al., 2018) deals with this by determining the density around a data point. If a point is surrounded by several other points, it is considered normal. However, if it is isolated from others, it is flagged as an outlier.

Another method called the Kernel Null Foley-Sammon Transform (KNFST) (Bodesheim et al., 2013) is an approach where data from known categories is presented into singular representative points. When new data is introduced, its difference from these points determines its novelty or how unusual it is. This means that if a new data point is far from any known category point, it is seen as more novel. Building on these methods, Liu et al. (2017) introduced the Incremental Kernel Null Space Based Discriminant Analysis (IKNDA). This method refines the process by analyzing the data, potentially making the detection of outliers more efficient and accurate.

Outliers are data points that do not follow the usual pattern. To find these outliers, some studies (Almohsen et al., 2022; Chen et al., 2023) use a method called self-representation, where data is divided into smaller groups or subspaces. Some studies (Hasan et al., 2016; Xu et al., 2017) use autoencoders to model what normal data looks like and then detect if any data that looks different.

GANs (Goodfellow et al., 2014) are also used by researchers for anomaly detection. For example, Wang et al. (2018) created new data that looks like the usual data. By doing this, they were able to measure how different each data point was from the usual pattern. Another study (Zhang et al., 2016) used GANs but focused on videos. A study conducted by Ionescu et al. (2019) trains a model to reconstruct noisy data by cleaning out the different ones, making the normal data stand out even more. In Sabokrou et al. (2018), a two phase framework was introduced for one-class classification and novelty detection. In first phase a network is trained to accurately reconstruct inliers. Whereas, in the second phase the other component is a one-class classifier, which utilizes a separate network to determine the novelty score of the data.

Recent studies (Li et al., 2021; Kendall and Gal, 2017; DeVries and Taylor, 2018; Liang et al., 2017) have worked into identifying out-of-distribution samples, essentially trying to find data that does not fit the usual distribution. They do this by finding the output entropy. One approach is to set a threshold for the softmax score, that helps in classification (Kendall and Gal, 2017). An advanced technique, known as ODIN (Liang et al., 2017), first changes the input data, which is adjusted through the gradients with respect to the input. Following this, ODIN combines the softmax score with scaling to enhance its detection capabilities.

While these strategies are promising, they rely on having labels for the regular, or in-distribution, data to help the training of the classifier networks. This could be a limitation in situations, where such labels might be sparse or entirely absent.

# 3 METHODOLOGY

In this section, we provide an overview of our network architecture and the training methodology employed to learn the mapping functions $f$ and $g$ as shown in Figure 2. These mappings, $g$ and $f$, are modeled using an autoencoder network. Our approach is based on previous novelty detection works (Almohsen et al., 2022; Pidhorskyi et al., 2018), which relies on the use of autoencoders (Rumelhart et al., 1986).

The primary objective in designing the autoencoder network and training procedure is to ensure that it captures the details of the underlying manifold $M$. For instance, if $M$ represents the distribution of images representing a specific object category, our aim is that the encoder and decoder should generate images that closely resemble the actual distribution. We introduce the latent space, denoted as $z$ (Pidhorskyi et al., 2018), which should closely match a predefined distribution, preferably a normal distribution with a mean of 0 and a standard deviation of 1, denoted as $N(0,1)$. Furthermore, we seek to make each component of $z$ maximally informative, so that they behave as independent random variables. This condition greatly facilitates the learning of a distribution $p_Z(z)$ from training data mapped onto the latent space $z$. Also, the autoencoder has generative capabilities, allowing us to generate data points $x \in M$ by sampling from $p_Z(z)$. This distinguishes our approach from Generative Adversarial Networks (GANs) (Goodfellow et al., 2014). Importantly, we also incorporate an encoder function $g$.

Variational Autoencoders (VAEs) (Kingma and Welling, 2013) have demonstrated their efficacy in handling continuous latent variables and generating data from a randomly sampled latent space. In contrast, Adversarial Autoencoders (AAEs) (Rumelhart et al., 1986) utilize an adversarial training paradigm to align the following distribution of the latent space with a specified distribution. A notable advantage of AAEs over VAEs is their ability to make the encoder match the entire prior distribution.

When dealing with images, both AAEs and VAEs often generate samples that deviate from the actual manifold (Pidhorskyi et al., 2018). This occurs because the decoder is updated based on a reconstruction loss, typically calculated as pixel-wise cross-entropy between the input and output images. This loss function tends to produce blur images, which can have adverse effects on our proposed approach. Similar to AAEs, PixelGAN autoencoders (Makhzani and Frey, 2017) also introduce an additional adversarial component in order to impose the prior distribution on latent space.

Following the methodology of previous studies (Pidhorskyi et al., 2018; Sabokrou et al., 2018, 2017), we introduce an additional adversarial training criterion that compares the output of the decoder network with the distribution of real data. It reduces the blurriness and enhances the local details of the generated images. We also introduce the different loss functions that calculate the losses for all the network components.

Our complete objective consists of three key components. Firstly, we calculate an adversarial loss through $Q_z$ discriminator, to match the latent space distribution with the predefined prior distribution, typically a standard normal distribution with a mean of 0 and a standard deviation of 1, denoted as $N(0,1)$. Secondly, adversarial loss from the distribution of decoded images is obtained from $z$ with the already-known distribution of training data i.e., inliers. Finally, autoencoder loss is used to quantify the dissimilarity between the decoded images and the original encoded input image using $Q_z$ discriminator. Figure 2 provides an illustration of our network configuration.

## 3.1 Adversarial losses

For the discriminator $Q_z$, the adversarial loss is defined as:

$$\mathcal{L}_{\text{adv-qz}}(x,g,Q_z) = \mathbb{E}\left[\log\left(Q_z(\mathcal{N}(0,1))\right)\right] + \mathbb{E}\left[\log\left(1 - Q_z(g(x))\right)\right] \qquad (1)$$

The objective of the encoder $g$ is to represent $x$ in $z$ in such a way that its distribution closely mirrors $N(0,1)$. $Q_z$ purpose is to differentiate between the encodings created by $g$ and a standard normal distribution. In this adversarial setting, while $g$ aims to reduce the value of this objective, $Q_z$ enhances it. Likewise, the adversarial loss associated with the discriminator $Q_y$ is:

$$\mathcal{L}_{\text{adv-qy}}(x,Q_y,m) = \mathbb{E}\left[\log\left(Q_y(x)\right)\right] + \mathbb{E}\left[\log\left(1 - Q_y(m(\mathcal{N}(0,1)))\right)\right] \qquad (2)$$

Table 1: Comparison of *F1* scores with prior research on MNIST (LeCun, 1998). The inliers consist of images from a single category, while outliers are randomly selected from the remaining categories. Our results are shown in bold outperforming previous works.

| % of outliers | $\mathcal{D}(R(X))$ (Sabokrou et al., 2018) | $\mathcal{D}(X)$ (Sabokrou et al., 2018) | LOF (Xia et al., 2015) | DRAE (You et al., 2017) | GPND (Base) (Pidhorskyi et al., 2018) | BK-AAND (Ours) |
|---|---|---|---|---|---|---|
| 10 | 0.97 | 0.93 | 0.92 | 0.95 | 0.98 | **0.989** |
| 20 | 0.92 | 0.90 | 0.83 | 0.91 | 0.97 | **0.985** |
| 30 | 0.92 | 0.87 | 0.72 | 0.88 | 0.96 | **0.980** |
| 40 | 0.91 | 0.84 | 0.65 | 0.82 | 0.95 | **0.976** |
| 50 | 0.88 | 0.82 | 0.55 | 0.73 | 0.94 | **0.972** |

Here the decoder *m* has the goal of generating *x* from a standard normal distribution, $N(0,1)$, such that *x* appears as though it was drawn from the original distribution. $Q_y$ role is to observe between the data points reconstructed by *m* and original data points *x*. In this way, *m* strives to reduce the objective function, whereas $Q_y$ works to increase it.

### 3.2 Autoencoder loss

For the combined optimization of encoder *g* and decoder *m*, the goal is to reduce the error for the input *x* that originates from an already-known data distribution. The error is defined as:

$$\mathcal{L}_{\text{error}}(x,g,m) = -\mathbb{E}_z[\log(p(m(g(x))|x))] \tag{3}$$

Here, $\mathcal{L}_{\text{error}}$ represents the negative expected log-likelihood, commonly known as the reconstruction error. Even though this loss lacks an adversarial aspect, it is used for autoencoder training. By reducing this loss, the functions *g* and *m* are considered accurate for the approximation of the original data.

### 3.3 Combined Loss Function

By combining all the previously mentioned loss functions, the total loss is formulated as:

$$\mathcal{L}(x,g,Q_z,Q_y,m) = \mathcal{L}_{\text{adv-qz}}(x,g,Q_z) + \mathcal{L}_{\text{adv-qy}}(x,Q_y,m) + \beta \mathcal{L}_{\text{error}}(x,g,m) \tag{4}$$

Here, β is a tuning parameter, that makes the relationship between the reconstruction loss and other losses. The autoencoder's optimal configuration is derived by minimizing the above equation, which yields:

$$(g^*, m^*) = \arg\min_{g,m} \max_{Q_y,Q_z} \mathcal{L}(x,g,Q_z,Q_y,m) \tag{5}$$

For the training process via stochastic gradient descent, each component is alternately updated:

- $Q_y$ weights are updated to maximize $\mathcal{L}_{\text{adv-qy}}$
- *m* weights are optimized to minimize $\mathcal{L}_{\text{adv-qy}}$
- $Q_z$ weights are updated to maximize $\mathcal{L}_{\text{adv-qz}}$
- Both *g* and *m* weights are optimized to minimize $\mathcal{L}_{\text{error}}$ and $\mathcal{L}_{\text{adv-qz}}$

## 4 EXPERIMENTS

Here, we present the assessment of our BK-AAND framework using three distinct datasets. The selection of these datasets is motivated by literature. We also provide an in-depth performance analysis and compare it to leading-edge methods. Moreover, we offer comprehensive discussions to demonstrate the robustness and importance of our proposed approach.

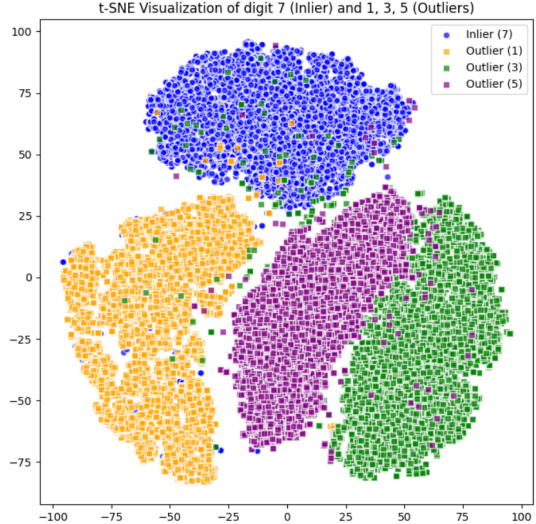

Figure 4: The t-SNE visualization of inlier (digit 7) and outliers (digits 1, 3, 5) classes. Each point represents a digit, and the colors distinguish between inliers and outliers. The visualization shows the distinct clusters formed in reduced two-dimensional space, offering insights into their latent representations. This visualization also shows a clear difference in the distribution to validate our experiments.

### 4.1 Datasets

We conducted our experiments using three widely used datasets: MNIST (LeCun, 1998), Coil-100 (Nene et al., 1996), and Fashion MNIST (Xiao et al., 2017). They are represented by *D1,D2* and *D3* in Figure 2.

**MNIST.** This dataset features handwritten digits from 0 to 9. Which is made up of 60,000 training images and 10,000 test images. These grayscale images have a size of 28x28 pixels. In our experiments, each of the ten categories is treated as inlier, while all other categories act as outliers.

**Coil-100.** This dataset captures 100 objects with 72 images per object, totaling 7,200 images. For our tests, we randomly choose $n$ categories $n \in \{1, 4, 7\}$ (Almohsen et al., 2022; Pidhorskyi et al., 2018) as inliers, and the remaining categories are treated as outliers.

**Fashion-MNIST.** This is a dataset of similar size, scale and format to MNIST, but for a different domain. The dataset contains 60,000 training images and 10,000 test images. The images are sized at 28x28 pixels in grayscale, and are in 10 classes corresponding to different fashion item categories (types of clothing/footwear).

### 4.2 Evaluation Metrics

We assess the effectiveness of our novelty detection methodology using two metrics, the $F_1$ score and the Area Under the Receiver Operating Characteristic curve (AUROC), which are widely used in related work (Chen et al., 2023; Li et al., 2021; Almohsen et al., 2022; Wang et al., 2022; Angiulli et al., 2020). All results presented can be requested from our implementation[1], which is coded in the advanced deep learning framework, PyTorch (Paszke et al., 2017).

### 4.3 Parameters and Implementation details

For the execution of experiments, we used two different machines. The first machine is a MacBook Pro 2023 with an M2 processor, equipped with an 8-core CPU, 24 GB of RAM, a 10-core GPU, and a 16-core neural engine specifically designed for deep learning computations. The second machine is a desktop with an NVIDIA GeForce 2080 Ti GPU and 24 GB of RAM.

---

[1] https://github.com/asad-python/BK-AAND

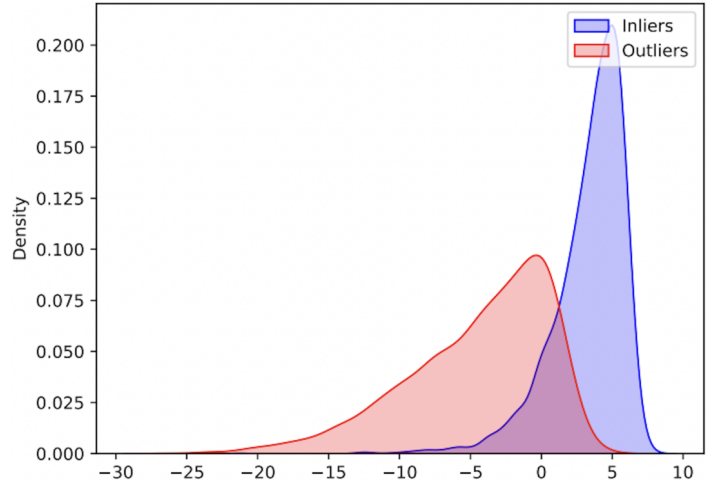

Figure 5: This image illustrates a Probability Density Function (PDF) plot on the MNIST dataset, providing a visual representation of the distribution of data. The peaks in the plot indicate the likelihood of different values occurring in the dataset. Inliers (seven categories of images), representing typical or expected data points, are depicted by the main body of the distribution, while outliers, which deviate significantly from the norm, stand out as distinct points in separate peaks.

Table 2: Results on Coil-100 dataset, inliers consist of images randomly selected from the 1,4 or 7 categories, while the outliers are chosen randomly from categories not included in the set of inliers, at most one outlier from each category

| Methods | **Inliers**: One category of images, **Outliers**: 50% | | **Inliers**: Four category of images, **Outliers**: 25% | | **Inliers**: Seven category of images, **Outliers**: 15% | |
| --- | --- | --- | --- | --- | --- | --- |
| | AUC | F1 | AUC | F1 | AUC | F1 |
| OutRank (Moonesinghe and Tan, 2008) | 0.836 | 0.862 | 0.613 | 0.491 | 0.570 | 0.342 |
| CoP (Rahmani and Atia, 2017) | 0.843 | 0.866 | 0.628 | 0.500 | 0.580 | 0.346 |
| REAPER (Lerman et al., 2015) | 0.900 | 0.892 | 0.877 | 0.703 | 0.824 | 0.541 |
| OutlierPursuit (Xu et al., 2010) | 0.908 | 0.902 | 0.837 | 0.686 | 0.822 | 0.528 |
| LRR (Liu et al., 2010) | 0.847 | 0.872 | 0.687 | 0.541 | 0.628 | 0.366 |
| DPCP (Tsakiris and Vidal, 2015) | 0.900 | 0.882 | 0.859 | 0.684 | 0.804 | 0.511 |
| $\ell_1$thresholding (Soltanolkotabi and Candes, 2012) | 0.991 | 0.978 | 0.992 | 0.941 | 0.991 | 0.897 |
| R-graph (You et al., 2017) | 0.997 | **0.990** | 0.996 | 0.970 | **0.996** | **0.955** |
| GPND (Pidhorskyi et al., 2018) | 0.968 | 0.979 | 0.945 | 0.960 | 0.919 | 0.941 |
| **BK-AAND (Ours)** | **0.998** | 0.957 | **0.997** | **0.972** | 0.844 | 0.929 |

Datasets were consistently partitioned into 5-folds. We adopted a base learning rate of 0.002, with 80 epochs. For our experiments, a batch size of 128 was chosen, while the latent size was set at 16. To rigorously evaluate the robustness of our method, we introduced varying outlier percentages to the dataset: 10%, 20%, 30%, 40%, and 50%. Specifically for the Coil-100 dataset, we further extended our outlier tests to include 15% and 25% in order to align our findings with previous research for comparative validity.

Other hyper-parameter details were kept the same as in Pidhorskyi et al. (2018). To ensure a fair and consistent comparative analysis, hyperparameters were kept the same as per specifications in related works.

## 5 RESULTS

We have evaluated our framework on three different datasets, varying the percentages of outliers. This section provides a detailed discussion on the results.

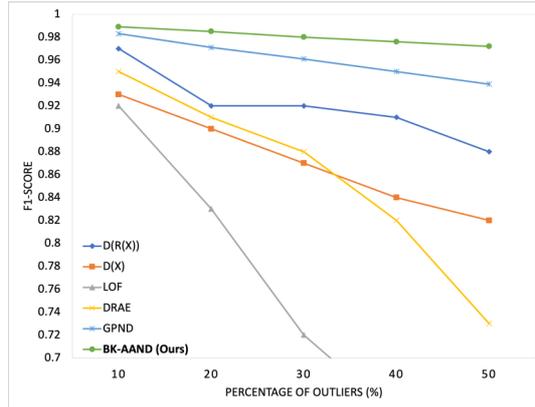

Figure 6: Comparison of results on MNIST (LeCun, 1998) dataset with previous literature. Images from a single category serve as inliers, while outliers are randomly selected from different categories.

Table 3: Fashion-MNIST (Xiao et al., 2017) results, where inliers are from the images of a single category, and we chose outliers from other random categories.

| % of Outliers | 10 | 20 | 30 | 40 | 50 |
|---|---|---|---|---|---|
| AUC | 0.968 | 0.937 | 0.942 | 0.937 | 0.962 |
| F1 | 0.97 | 0.942 | 0.918 | 0.889 | 0.927 |

## 5.1 MNIST dataset

We adopt a methodology close to the one described in Sabokrou et al. (2018) and Pidhorskyi et al. (2018), adding several distinct modifications. Our outcomes derive from the mean of 5-fold cross-validation, wherein each fold has 20% of every class. For every digit class, our model undergoes training, with outliers introduced by randomly selecting images from alternate categories, their proportion varying between 10% to 50%. Compared to the approach in Sabokrou et al. (2018), where data is not segmented into discrete training, validation, and testing sets, we take a different approach. Specifically, we avoid using the same inliers for both training and testing, fixing to a 60%, 20%, and 20% division for the training, validation, and test sets. The results on the MNIST dataset are shown in Table 1, comparison with other approaches (Breunig et al., 2018; Sabokrou et al., 2018; Pidhorskyi et al., 2018; Xia et al., 2015) in Figure 6. Reconstruction of inliers and outliers is visually represented in Figure 1 for better comparison. Results show that our approach is performing better than previous studies when tested on different percentages of outliers. The t-SNE visualization in Figure 4 and PDF plot in Figure 5 is also illustrated to validate the distribution of the data and predictions.

## 5.2 Coil-100 dataset

We followed the method in Pidhorskyi et al. (2018) with a few changes related to the percentages of the outliers. Our results are an average from the 5-fold cross-validation. In this process, 20% of each group (class) is tested. Since there are not many samples in each group, we train with 80% and test with 20%. The study in You et al. (2017) did not separate their data for training, validating, and testing. They did not need to because they used a special network called VGG (Simonyan and Zisserman, 2014) that was already trained with ImageNet (Russakovsky et al., 2015) as discussed in Pidhorskyi et al. (2018). Unlike them, we did not reuse some parts (inliers) and stuck to using 80% for training and 20% for testing.

Reconstruction of outliers is shown in Figure 3. Our results for Coil-100 are in Table 2. We did not do better than R-graph (You et al., 2017) in a few cases i.e.,15% of outliers. But it is important to remember that R-graph worked with a VGG network that was already trained. We started from scratch and trained a system (BK-AAND) using a very small amount of data i.e., 70 samples for each group.

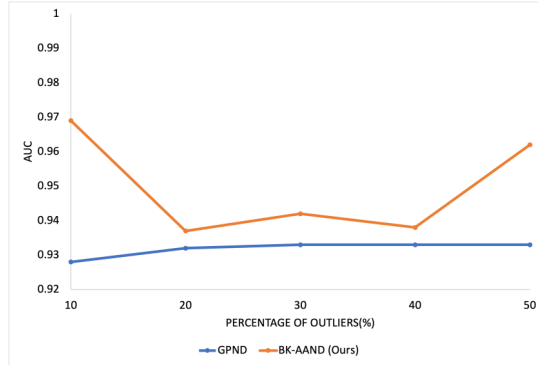

Figure 7: Comparison of results on Fashion-MNIST with GPND (Pidhorskyi et al., 2018). Single categories of images are considered at inliers and for outliers, we randomly choose the categories as in GPND.

### 5.3 Fashion-MNIST dataset

For this dataset, we have repeated the same experiments with the same protocol that is used for MNIST. Please refer to 5.1 for the experimental setup. Table 3 shows the results achieved on this Fashion-MNIST. Moreover, the comparison with the base GPND paper (Pidhorskyi et al., 2018) is also provided in Figure 7. We have compared GPND on different numbers of percentages and it can be seen that it outperforms the base model on all those percentages of outliers.

## 6 CONCLUSIONS

### 6.1 Key Findings

We proposed BK-AAND, an approach along with a network structure for novelty detection. This is designed around deriving mappings $f$ and $g$ which presents the parameterized manifold $M$. This manifold captures the structure of the inlier distribution. Compared with existing deep learning methodologies, BK-AAND identifies outliers by considering their inlier probability distribution and reconstruction loss. Moreover, despite its simple design, our approach has been demonstrated to deliver outperforming results across different metrics, datasets, and protocols. The outcomes from multiple datasets indicate that our suggested approach can identify samples outside the target class (meaning they're novel, outliers, or anomalies), despite the absence of such samples in the training phase. Through extensive testing, we found that our method consistently produces reliable outcomes over numerous training epochs and outperforms various state-of-the-art techniques (Almohsen et al., 2022; Chen et al., 2023; Pidhorskyi et al., 2018; Sabokrou et al., 2018; Xia et al., 2015; You et al., 2017; Xu et al., 2010; Moonesinghe and Tan, 2008; Lerman et al., 2015; Rahmani and Atia, 2017; Liu et al., 2010; Breunig et al., 2018; Tsakiris and Vidal, 2015; Soltanolkotabi and Candes, 2012), in detecting novelties. Moreover, our model is lightweight because it has fewer parameters, a simple shallow architecture, and low memory needs.

### 6.2 Future Directions

In future work, we hope to train and evaluate our method on large scale datasets. We also intend to perform a statistical comparison of the execution time to support the efficiency of our lightweight model.

# ACKNOWLEDGEMENTS

This research is conducted with the financial support of Science Foundation Ireland under Grant number 18/CRT/6223, partnered with Valeo.

# References

Almohsen, R., Keaton, M. R., Adjeroh, D. A., and Doretto, G. (2022). Generative probabilistic novelty detection with isometric adversarial autoencoders. In *Proceedings of the IEEE/CVF Conference on Computer Vision and Pattern Recognition*, pages 2003–2013.

Angiulli, F., Basta, S., Lodi, S., and Sartori, C. (2020). Reducing distance computations for distance-based outliers. *Expert Systems with Applications*, 147:113215.

Bergadano, F. (2019). Keyed learning: An adversarial learning framework—formalization, challenges, and anomaly detection applications. *ETRI Journal*, 41(5):608–618.

Bodesheim, P., Freytag, A., Rodner, E., Kemmler, M., and Denzler, J. (2013). Kernel null space methods for novelty detection. In *Proceedings of the IEEE conference on computer vision and pattern recognition*, pages 3374–3381.

Breunig, M. M., Kriegel, H.-P., Ng, R. T., and Sander, J. (2018). Lof: identifying density-based local outliers. In *Proceedings of the 2000 ACM SIGMOD international conference on Management of data*, pages 93–104.

Chen, Y., Cheng, L., Hua, Z., and Yi, S. (2023). Laplacian regularized deep low-rank subspace clustering network. *Applied Intelligence*, pages 1–15.

DeVries, T. and Taylor, G. W. (2018). Learning confidence for out-of-distribution detection in neural networks. *arXiv preprint arXiv:1802.04865*.

Gong, D., Liu, L., Le, V., Saha, B., Mansour, M. R., Venkatesh, S., and Hengel, A. v. d. (2019). Memorizing normality to detect anomaly: Memory-augmented deep autoencoder for unsupervised anomaly detection. In *Proceedings of the IEEE/CVF International Conference on Computer Vision*, pages 1705–1714.

Goodfellow, I., Pouget-Abadie, J., Mirza, M., Xu, B., Warde-Farley, D., Ozair, S., Courville, A., and Bengio, Y. (2014). Generative adversarial nets. *Advances in neural information processing systems*, 27.

Hasan, M., Choi, J., Neumann, J., Roy-Chowdhury, A. K., and Davis, L. S. (2016). Learning temporal regularity in video sequences. In *Proceedings of the IEEE conference on computer vision and pattern recognition*, pages 733–742.

Hinami, R., Mei, T., and Satoh, S. (2017). Joint detection and recounting of abnormal events by learning deep generic knowledge. In *Proceedings of the IEEE international conference on computer vision*, pages 3619–3627.

Humbert, P., Le Bars, B., and Minvielle, L. (2022). Robust kernel density estimation with median-of-means principle. In *International Conference on Machine Learning*, pages 9444–9465. PMLR.

Ionescu, R. T., Khan, F. S., Georgescu, M.-I., and Shao, L. (2019). Object-centric auto-encoders and dummy anomalies for abnormal event detection in video. In *Proceedings of the IEEE/CVF Conference on Computer Vision and Pattern Recognition*, pages 7842–7851.

Jiang, X., Liu, J., Wang, J., Nie, Q., Wu, K., Liu, Y., Wang, C., and Zheng, F. (2022). Softpatch: Unsupervised anomaly detection with noisy data. *Advances in Neural Information Processing Systems*, 35:15433–15445.

Kendall, A. and Gal, Y. (2017). What uncertainties do we need in bayesian deep learning for computer vision? *Advances in neural information processing systems*, 30.

Kingma, D. P. and Welling, M. (2013). Auto-encoding variational bayes. *arXiv preprint arXiv:1312.6114*.

LeCun, Y. (1998). The mnist database of handwritten digits. *http://yann. lecun. com/exdb/mnist/*.

Lerman, G., McCoy, M. B., Tropp, J. A., and Zhang, T. (2015). Robust computation of linear models by convex relaxation. *Foundations of Computational Mathematics*, 15:363–410.

Li, W., Huang, X., Lu, J., Feng, J., and Zhou, J. (2021). Learning probabilistic ordinal embeddings for uncertainty-aware regression. In *Proceedings of the IEEE/CVF conference on computer vision and pattern recognition*, pages 13896–13905.

Liang, S., Li, Y., and Srikant, R. (2017). Enhancing the reliability of out-of-distribution image detection in neural networks. *arXiv preprint arXiv:1706.02690*.

Liu, G., Lin, Z., and Yu, Y. (2010). Robust subspace segmentation by low-rank representation. In *Proceedings of the 27th international conference on machine learning (ICML-10)*, pages 663–670.

Liu, J., Lian, Z., Wang, Y., and Xiao, J. (2017). Incremental kernel null space discriminant analysis for novelty detection. In *Proceedings of the IEEE Conference on Computer Vision and Pattern Recognition*, pages 792–800.

Liu, W., Luo, W., Lian, D., and Gao, S. (2018). Future frame prediction for anomaly detection–a new baseline. In *Proceedings of the IEEE conference on computer vision and pattern recognition*, pages 6536–6545.

Liu, Z., Zhou, Y., Xu, Y., and Wang, Z. (2023). Simplenet: A simple network for image anomaly detection and localization. In *Proceedings of the IEEE/CVF Conference on Computer Vision and Pattern Recognition*, pages 20402–20411.

Luo, W., Liu, W., and Gao, S. (2017). A revisit of sparse coding based anomaly detection in stacked rnn framework. In *Proceedings of the IEEE international conference on computer vision*, pages 341–349.

Makhzani, A. and Frey, B. J. (2017). Pixelgan autoencoders. *Advances in Neural Information Processing Systems*, 30.

Moonesinghe, H. and Tan, P.-N. (2008). Outrank: a graph-based outlier detection framework using random walk. *International Journal on Artificial Intelligence Tools*, 17(01):19–36.

Nene, S. A., Nayar, S. K., Murase, H., et al. (1996). Columbia object image library (coil-20).

Paszke, A., Gross, S., Chintala, S., Chanan, G., Yang, E., DeVito, Z., Lin, Z., Desmaison, A., Antiga, L., and Lerer, A. (2017). Automatic differentiation in pytorch.

Pidhorskyi, S., Almohsen, R., and Doretto, G. (2018). Generative probabilistic novelty detection with adversarial autoencoders. *Advances in neural information processing systems*, 31.

Rahmani, M. and Atia, G. K. (2017). Coherence pursuit: Fast, simple, and robust principal component analysis. *IEEE Transactions on Signal Processing*, 65(23):6260–6275.

Ravanbakhsh, M., Nabi, M., Mousavi, H., Sangineto, E., and Sebe, N. (2018). Plug-and-play cnn for crowd motion analysis: An application in abnormal event detection. In *2018 IEEE Winter Conference on Applications of Computer Vision (WACV)*, pages 1689–1698. IEEE.

Ravanbakhsh, M., Nabi, M., Sangineto, E., Marcenaro, L., Regazzoni, C., and Sebe, N. (2017). Abnormal event detection in videos using generative adversarial nets. In *2017 IEEE international conference on image processing (ICIP)*, pages 1577–1581. IEEE.

Ren, H., Liu, W., Olsen, S. I., Escalera, S., and Moeslund, T. B. (2015). Unsupervised behavior-specific dictionary learning for abnormal event detection. In *British Machine Vision Conference 2015: Machine Vision of Animals and their Behaviour*, pages 28–1. British Machine Vision Association.

Rumelhart, D. E., Hinton, G. E., and Williams, R. J. (1986). Learning representations by back-propagating errors. *nature*, 323(6088):533–536.

Russakovsky, O., Deng, J., Su, H., Krause, J., Satheesh, S., Ma, S., Huang, Z., Karpathy, A., Khosla, A., Bernstein, M., et al. (2015). Imagenet large scale visual recognition challenge. *International journal of computer vision*, 115:211–252.

Sabokrou, M., Fayyaz, M., Fathy, M., and Klette, R. (2017). Deep-cascade: Cascading 3d deep neural networks for fast anomaly detection and localization in crowded scenes. *IEEE Transactions on Image Processing*, 26(4):1992–2004.

Sabokrou, M., Khalooei, M., Fathy, M., and Adeli, E. (2018). Adversarially learned one-class classifier for novelty detection. In *Proceedings of the IEEE conference on computer vision and pattern recognition*, pages 3379–3388.

Simonyan, K. and Zisserman, A. (2014). Very deep convolutional networks for large-scale image recognition. *arXiv preprint arXiv:1409.1556*.

Smeureanu, S., Ionescu, R. T., Popescu, M., and Alexe, B. (2017). Deep appearance features for abnormal behavior detection in video. In *Image Analysis and Processing-ICIAP 2017: 19th International Conference, Catania, Italy, September 11-15, 2017, Proceedings, Part II 19*, pages 779–789. Springer.

Soltanolkotabi, M. and Candes, E. J. (2012). A geometric analysis of subspace clustering with outliers.

Sultani, W., Chen, C., and Shah, M. (2018). Real-world anomaly detection in surveillance videos. In *Proceedings of the IEEE conference on computer vision and pattern recognition*, pages 6479–6488.

Tsakiris, M. C. and Vidal, R. (2015). Dual principal component pursuit. In *Proceedings of the IEEE International Conference on Computer Vision Workshops*, pages 10–18.

Wang, H.-g., Li, X., and Zhang, T. (2018). Generative adversarial network based novelty detection using minimized reconstruction error. *Frontiers of Information Technology & Electronic Engineering*, 19:116–125.

Wang, S., Zeng, Y., Yu, G., Cheng, Z., Liu, X., Zhou, S., Zhu, E., Kloft, M., Yin, J., and Liao, Q. (2022). E$\hat{\ }\{3\}$3 outlier: a self-supervised framework for unsupervised deep outlier detection. *IEEE Transactions on Pattern Analysis and Machine Intelligence*, 45(3):2952–2969.

Xia, Y., Cao, X., Wen, F., Hua, G., and Sun, J. (2015). Learning discriminative reconstructions for unsupervised outlier removal. In *Proceedings of the IEEE international conference on computer vision*, pages 1511–1519.

Xiao, H., Rasul, K., and Vollgraf, R. (2017). Fashion-mnist: a novel image dataset for benchmarking machine learning algorithms. *arXiv preprint arXiv:1708.07747*.

Xu, D., Ricci, E., Yan, Y., Song, J., and Sebe, N. (2015). Learning deep representations of appearance and motion for anomalous event detection. *arXiv preprint arXiv:1510.01553*.

Xu, D., Yan, Y., Ricci, E., and Sebe, N. (2017). Detecting anomalous events in videos by learning deep representations of appearance and motion. *Computer Vision and Image Understanding*, 156:117–127.

Xu, H., Caramanis, C., and Sanghavi, S. (2010). Robust pca via outlier pursuit. *Advances in neural information processing systems*, 23.

Xu, H., Zhang, L., Li, P., and Zhu, F. (2022). Outlier detection algorithm based on k-nearest neighbors-local outlier factor. *Journal of Algorithms & Computational Technology*, 16:17483026221078111.

Yamanishi, K., Takeuchi, J.-I., Williams, G., and Milne, P. (2000). On-line unsupervised outlier detection using finite mixtures with discounting learning algorithms. In *Proceedings of the sixth ACM SIGKDD international conference on Knowledge discovery and data mining*, pages 320–324.

You, C., Robinson, D. P., and Vidal, R. (2017). Provable self-representation based outlier detection in a union of subspaces. In *Proceedings of the IEEE conference on computer vision and pattern recognition*, pages 3395–3404.

Zhang, Y., Lu, H., Zhang, L., Ruan, X., and Sakai, S. (2016). Video anomaly detection based on locality sensitive hashing filters. *Pattern Recognition*, 59:302–311.

Zhou, J. and Xing, H. J. (2023). Novelty detection method based on dual autoencoders and transformer network. *Journal of Computer Applications*, 43(1):22.

Zhou, Y. (2022). Rethinking reconstruction autoencoder-based out-of-distribution detection. In *Proceedings of the IEEE/CVF Conference on Computer Vision and Pattern Recognition*, pages 7379–7387.